\documentclass[11pt]{article}

\usepackage{acl}

\usepackage{times}
\usepackage{latexsym}
\usepackage{booktabs}
\usepackage{caption}
\usepackage{graphicx}

\usepackage[T1]{fontenc}
\usepackage[utf8]{inputenc}

\usepackage{microtype}

\usepackage{inconsolata}

\usepackage{graphicx}

\title{MedAI: Evaluating TxAgent's Therapeutic Agentic Reasoning in the NeurIPS CURE-Bench Competition}

\author{
    Tim Cofala \\
    L3S Research Center \\
    tim.cofala@l3s.de
    \And
    Christian Kalfar \\
    L3S Research Center \\
    christian.kalfar@l3s.de
    \And
    Jingge Xiao \\
    L3S Research Center \\
    xiao@l3s.de
    \AND
    Johanna Schrader \\
    L3S Research Center \\
    schrader@l3s.de
    \And
    Michelle Tang \\
    L3S Research Center \\
    tang@l3s.de
    \And
    Wolfgang Nejdl \\
    L3S Research Center \\
    nejdl@l3s.de
}

\begin{document}
\maketitle

\begin{abstract}
Therapeutic decision-making in clinical medicine constitutes a high-stakes domain in which AI guidance interacts with complex interactions among patient characteristics, disease processes, and pharmacological agents. Tasks such as drug recommendation, treatment planning, and adverse-effect prediction demand robust, multi-step reasoning grounded in reliable biomedical knowledge. Agentic AI methods, exemplified by TxAgent, address these challenges through iterative retrieval-augmented generation (RAG). TxAgent employs a fine-tuned Llama-3.1-8B model that dynamically generates and executes function calls to a unified biomedical tool suite (ToolUniverse), integrating FDA Drug API, OpenTargets, and Monarch resources to ensure access to current therapeutic information. In contrast to general-purpose RAG systems, medical applications impose stringent safety constraints, rendering the accuracy of both the reasoning trace and the sequence of tool invocations critical. These considerations motivate evaluation protocols treating token-level reasoning and tool-usage behaviors as explicit supervision signals. This work presents insights derived from our participation in the CURE-Bench NeurIPS 2025 Challenge, which benchmarks therapeutic-reasoning systems using metrics that assess correctness, tool utilization, and reasoning quality. We analyze how retrieval quality for function (tool) calls influences overall model performance and demonstrate performance gains achieved through improved tool-retrieval strategies. Our work was awarded the \textit{Excellence Award in Open Science}. Complete information can be found at \href{https://curebench.ai/}{curebench.ai.}
\end{abstract}
\section{Introduction}
% Motivate clinical decision-making
% Explain the difficulty of the task
% Highlight limitations (multi-step reasoning, correct tool calling, token usage, ...)
% Introduce Curebench (2 tracks)
% Why is tool calling important? Justification for track 2
% TxAgent already good + brief description (+ Tool Universe)
% Very brief description of what we did and what we achieved

Therapeutic decision-making in clinical medicine presents a demanding environment for artificial intelligence. Clinicians routinely integrate heterogeneous information on patient characteristics, disease pathophysiology, comorbid conditions and the pharmacological properties of candidate treatments. AI systems intended to support such decisions must therefore demonstrate not only competent prediction capabilities but also the capacity to reason through multi-step therapeutic processes in a manner that is grounded in reliable biomedical knowledge.

Recent advances in agentic AI \cite{MedAgents, ChemCrow, ReACT} and retrieval-augmented generation (RAG) \cite{instructRAG, dynamicRAG, AstuteRAG} have introduced new opportunities for building systems that can navigate complex biomedical toolchains. Rather than relying solely on parametric knowledge, agentic approaches iteratively retrieve, evaluate, and integrate external information sources through orchestrated tool use. These methods are promising for therapeutic applications, where accurate access to up-to-date drug and disease information is necessary for safe model operation \cite{TxAgent}. However, general-purpose agentic frameworks are not by themselves sufficient: medical contexts impose stringent constraints on verifiability and errors in either the reasoning trace or the sequence of tool calls can propagate to clinically significant mistakes \cite{gorenshtein2025ai, ClinicalSafetyFramework, MedPaLM, LLMsInMedicine}. As a result, evaluating therapeutic-reasoning systems requires protocols that directly assess reasoning quality, tool utilization, and the correctness of answers and intermediate steps.

The Agentic Tool-Augmented Reasoning track of the \href{https://curebench.ai/}{CURE-Bench NeurIPS 2025 Challenge} establishes a rigorous framework for evaluating these capabilities. By combining metrics for answer accuracy, tool utilization, and reasoning validity with expert human review, the challenge ensures that agentic systems for therapeutic reasoning are assessed with the necessary precision and care.

Building on the framework of this challenge, our effort focused on enhancing TxAgent \cite{TxAgent}, an agentic therapeutic-reasoning system built on a fine-tuned Llama-3.1-8B model equipped with the ToolUniverse \cite{ToolUniverse}, a unified suite of biomedical resources integrating FDA drug data, OpenTargets associations, and Monarch ontologies. 

We conducted a detailed analysis of TxAgent's performance on the CURE-Bench challenge, with a particular focus on retrieval quality for function (tool) calls, as retrieval failures were frequently responsible for downstream reasoning errors. 

To address these issues, we investigated approaches to improve the system by:
\begin{enumerate}
    \item Integrating DailyMed to access up-to-date drug label information; 
    \item Investigating the impact of fixed retrieval scenarios on the performance of vanilla LLMs in therapeutic decision-making; 
    \item Comparing state-of-the-art retrieval approaches with the TxAgent workflow, including integration with DailyMed.
\end{enumerate}

The integration of DailyMed significantly improved the system's access to up-to-date drug-label information. The resulting performance gains within the CURE-Bench framework contributed to our team receiving the Excellence Award in Open Science, highlighting the effectiveness of targeted improvements in retrieval and decision-making workflows.

\section{Related Work}
\label{sec:related-works}
In this section, we describe the context of tool-calling for context enhancement, and clarify how this process differs from retrieval-augmented generation.
%in the development of medical foundation models. 
\newline

\noindent \textbf{Retrieval-Augmented Generation Systems.}
Retrieval-augmented generation (RAG) systems may include various framework-specific components \cite{yang2024rag, ye2024boosting} but at least incorporate a retriever component \cite{asai2023self, hambarde2023information, yang2024rag}.
SotA LLM's factual errors \cite{mallen2023trustlanguagemodelsinvestigating, min2023factscorefinegrainedatomicevaluation} are reduced by enhancing the LLM prompt with additional query-based retrieved information, reducing hallucinations \cite{yang2024rag, asai2023self} through including up-to-date information \cite{jaenich2024fairness}.
Recent research demonstrates the impact of retrievers and functionalities on RAG systems, among others, that sparse retrievers excel in capturing lexical tasks while dense retrievers excel in semantic tasks \cite{cuconasu2024power, hambarde2023information}.
\newline

\noindent \textbf{Tool-Calling.}
While retrieval focuses on retrieving information from a fixed corpus, calling tools provide a framework that is used for specialized retrieval, which can additionally replace emerging capabilities of LLMs. 
Fundamental retrieval tasks, QA datasets based on Wikipedia corpus \cite{instructRAG}, have been extensively studied over the past years, focusing on increasing the precision and recall of retrieving ground-truth documents. Additional performance improvements have been achieved through RAG-solutions by including additional components to guide or judge the retrieval process through additional retrieval rounds \cite{yang2024rag}, self-reflection \cite{DBLP:journals/bioinformatics/JeongSSK24}, or adaptive retrieval \cite{zhang2024retrievalqaa, liu2024ctrlaadap, asai2023self}.
Calling tools, accessing up-to-date information \cite{TxAgent} or outsourcing capabilities \cite{DBLP:conf/nips/SchickDDRLHZCS23}, through API endpoints builds upon the RAG structure and previous works. In both scenarios, the task at hand requires being abstracted to a valid function call considering the intention and available functions \cite{DBLP:conf/nips/PatilZ0G24}. Additionally, an execution function retrieves and parses the information back into the LLM context \cite{TxAgent}.
\newline

%\noindent \textbf{Medical Foundation Models.}
%Medical foundation models focus on the medical domain and provide solutions ranging from protein data \cite{scGPT, scFoundation}, therapeutic reasoning \cite{TxAgent}, or structure a wide range of medical tasks \cite{Biomni}.

\section{TxAgent}\label{sec:rag-structure}

Agentic approaches extend LLMs with a surrounding framework guiding the LLM through a defined process. Here, we outline the key components of the agentic TxAgent \cite{TxAgent} and its workflow.

\subsection{TxAgent's Workflow}
TxAgent \cite{TxAgent} consists of two key components. The first is the finetuned Llama3.1-8B LLM, the second the finetuned Qwen2-1.5B. A therapeutic question regarding a patient's condition is the question at hand to be answered. A query similar to \textit{What are the side-effects to consider taking [drug] for a pregnant woman?} stands at the beginning of the iterative answer generation.

In the first step, the LLM reformulates the query to highlight the intention \textit{side-effects} by considering the stated intention. This statement is compared against all function call descriptions implemented in ToolUniverse \cite{ToolUniverse}. TxAgent relies on the ToolUniverse framework for information retrieval, which provides access to medical drug databases, such as the FDA and OpenTargets. Qwen2-1.5B returns the most promising function calls back to Llama3.1-8B for tool selection. In TxAgent, this process is called: \textit{ToolRAG}.
TxAgent continues with the selection of the returned tools. It independently decides which and how many function calls to perform, and how to construct them. After receiving function call suggestions, TxAgent defines the required parameters for each call. In a single \textit{ToolRAG} round, TxAgent can invoke multiple tool calls using different parameters. These combinations of function names and parameters are generated in JSON format, parsed by the framework, and then executed by ToolUniverse in an independent loop before the information is fed back to the LLM.

The retrieved information is incorporated back into the LLM, starting a new iteration. In each iteration, the LLM decides whether additional up-to-date information is needed to answer the question, based on previously retrieved information, generating \textit{ToolRAG} cycles. This need can arise from insufficient function calls, failed function calls, wrongly formatted calls, or insufficiently selected functions. Additionally, new information may require further retrieval operations for clarification, e.g., a specific drug is mentioned in the retrieved context. Otherwise, the framework terminates and returns a final statement. In RAG scenarios, the generated context is used to answer the questions, and the function call trace with its parameter is additionally returned for answer generation tracking.

\subsection{Focusing on Function Calls}

Our work focuses on a critical aspect of tool-calling: selecting the right tool. TxAgent \cite{TxAgent} uses a finetuned encoder–decoder pipeline in which Llama3.1-8B rewrites the question, while Qwen2-1.5B compares the rewritten question to ToolUniverse \cite{ToolUniverse} function descriptions and returns the top-$k=10$ tool names based on cosine similarity.

Analyzing the evaluation runs of function calls, we observe several recurring issues: functions are called repeatedly due to incorrectly formatted input parameter names; incorrect functions are selected even when better candidates are retrieved; and some function calls fail to return the expected information.

%To provide TxAgent with additional access to up-to-date drug-related information, we extend the ToolUniverse framework with another public drug database, DailyMed. FDA and DailyMed cover similar drug information. ToolUniverse is implemented to return only the specific dictionary-key context for FDA functions and a similar implementation would be possible for DailyMed. However, since DailyMed offers smaller context sections and does not cover all the information returned by the FDA endpoint, we instead choose to return the complete DailyMed context to the LLM. By enabling TxAgent to call the DailyMed function, we introduce a new API endpoint to ToolUniverse with tool descriptions analogous to those of the FDA and OpenTarget functions. Since the tool-selection cosine-similarity mechanism can now return DailyMed as a suggested function, TxAgent is able to call it using the parameter types defined within ToolUniverse. 
We enhanced TxAgent's pharmacological data capabilities by integrating the DailyMed database into the ToolUniverse framework. While TxAgent's existing openFDA tools facilitate granular queries and metadata retrieval, they often do so at the expense of contextual depth. Consequently, these tools cannot provide the comprehensive, human-readable clinical narratives required for robust medical reasoning in a single function call. Although the granular approach of openFDA optimizes token usage for specific queries, it necessitates multiple function calls to resolve broader therapeutic questions. Therefore, DailyMed and openFDA fulfill distinct but complementary roles within our approach: the DailyMed tool integration grants TxAgent direct access to authoritative Structured Product Labeling (SPL), ensuring the retrieval of complete, version-controlled clinical narratives. By registering a semantic description of our DailyMed tool within ToolUniverse, we enable TxAgent to autonomously discover and invoke this resource via the existing tool-selection mechanism.

\section{Experiments}

Our experiments investigate a fixed-retrieval setup, highlighting how the capabilities of LLMs influence overall performance. We further compare a range of sparse and dense retrieval systems for tool retrieval, alongside the performance of enhanced TxAgent retrieval with Dailymed. As our experiments rely on the provided datasets and require ground-truth labels, we first describe the dataset used.

\subsection{Dataset}
Throughout the competition, participants were presented with a total of three datasets, as listed in Table \ref{tab:1-datasets}: a \textit{validation} set containing 459 questions with ground-truth answers, and two \textit{test} sets containing 2,097 and 2,491 questions, respectively. The validation set was used to develop and implement solutions for the competition, while the test sets, used by the organizers to evaluate performance, do not contain ground-truth labels.

\begin{table}[ht]
\centering
\caption{Overview of the competition datasets, including question styles and question counts.}
\resizebox{0.49\textwidth}{!}{  % scale to 80% of text width
\begin{tabular}{lccc}
\toprule
Dataset (\#Qs) & MC [n] & OE-MC [n] & OE [n] \\ 
\midrule
Validation (459) & 183 (39.9\%) & 230 (50.1\%) & 46 (10.0\%)\\
Test 1 (2097) & 663 (31.6\%) & 1274 (60.8\%) & 142 (6.8\%) \\
Test 2 (2491) & 779 (31.3\%) & 1474 (59.2\%) & 238 (9.6\%) \\ 
\bottomrule
\end{tabular}
}
\label{tab:1-datasets}
\end{table}

All datasets include three different question styles:  
\begin{enumerate}
    \item \textbf{Open-ended (OE)}: A free-text answer is expected.  
    \item \textbf{Multiple-choice (MC)}: Four answer options are provided, and the LLM must select the correct one.  
    \item \textbf{Open-ended multiple-choice (OE-MC)}: A two-step approach where an open-ended answer is first generated and then used as context to select the correct option from multiple choices. 
\end{enumerate}

To maximize the number of questions with ground-truth labels, we generated multiple-choice and open-ended multiple-choice versions for each question. In the following, \textbf{MC} refers to the multiple-choice validation dataset, and \textbf{OE-MC} refers to the open-ended multiple-choice validation dataset.

\subsection{Retriever Comparison}

\begin{figure}[h]
    \caption{State-of-the-art dense retriever present similar performance but sparse retriever BM25. TxAgent's finetuned Qwen2-1.5B retriever outperforms all retriever and enhances with DailyMed.}
    \includegraphics[width=\columnwidth]{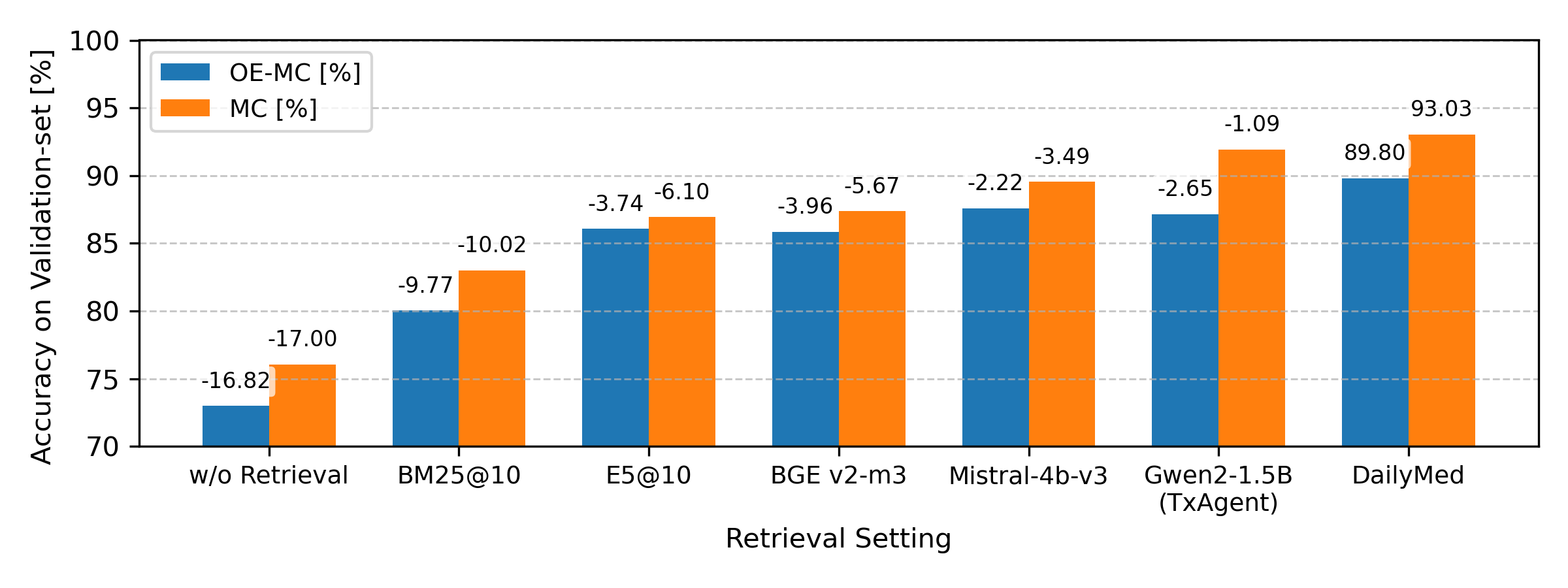}
    \label{fig:retriever}
\end{figure}

To evaluate the performance gains achieved by incorporating DailyMed into TxAgent, we compared the accuracy of several state-of-the-art sparse and dense retrievers. To this end, we rewrote TxAgent's ToolRAG functionality to support various retrievers. The retrieval decision process, however, was kept consistent with the original TxAgent workflow: each retriever compares the rewritten user question generated by TxAgent to the function descriptions in ToolUniverse, and only the top-$k = 10$ function names are returned.

Figure~\ref{fig:retriever} contains the overview of all 4 investigated sparse and dense retriever, as well as the performance of TxAgent without retrieval. The percentage values per bar chart mention the performance decrease relative to the highest performence setting of TxAgent with the integration of DailyMed.

No additional information impacts Llama3.1-8B, as this LLM cannot leverage extra medical knowledge necessary for the tasks at hand. The sparse retriever BM25 struggles to retrieve the correct function names due to its reliance on exact word matches and the limited context of function descriptions, which consist of at most two sentences per function in ToolUniverse. Dense retrievers perform similarly, albeit with higher runtime variability. However, the total number of tokens to rerank remains negligible, since each function is described by only up to two sentences.

TxAgent itself uses a dense retriever with Qwen2-1.5B, which has been further fine-tuned on therapeutic medical questions with additional ground-truth tool-call labels that were not available to competition participants. The performance of Qwen2-1.5B is only surpassed when integrating DailyMed, whose one-sentence function descriptions provide sufficient context for Qwen2-1.5B to correctly select the DailyMed function.

\subsection{No-RAG, RAG, and TxAgent}

In this section, we describe the experimental setup for the fixed retrieval setting and provide a detailed analysis of the results.

\begin{figure}[h]
    \caption{Comparison of frozen retrieval performance across various LLMs, measured by accuracy [\%] on OE-MC questions in both RAG and no-retrieval settings, with optional permuted multiple-choice answer options, under the baseline performance of TxAgent without modification.}
    \includegraphics[width=\columnwidth]{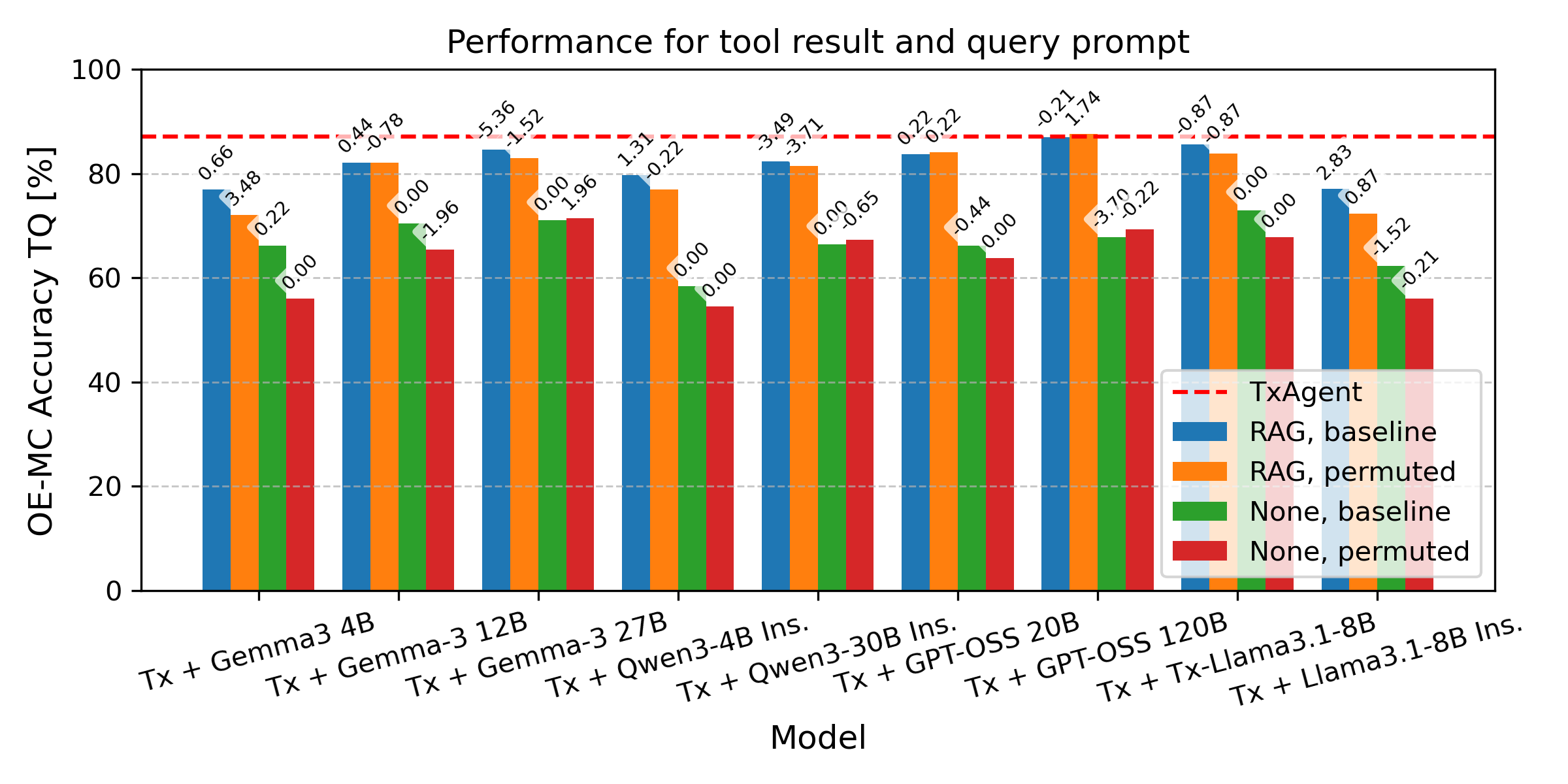}
    \label{fig:acc_oe}
\end{figure}

\begin{figure}[h]
    \caption{Comparison of frozen retrieval performance across various LLMs, measured by accuracy [\%] on MC questions in both RAG and no-retrieval settings, with optional permuted multiple-choice answer options, under the baseline performance of TxAgent without modification.}
    \includegraphics[width=\columnwidth]{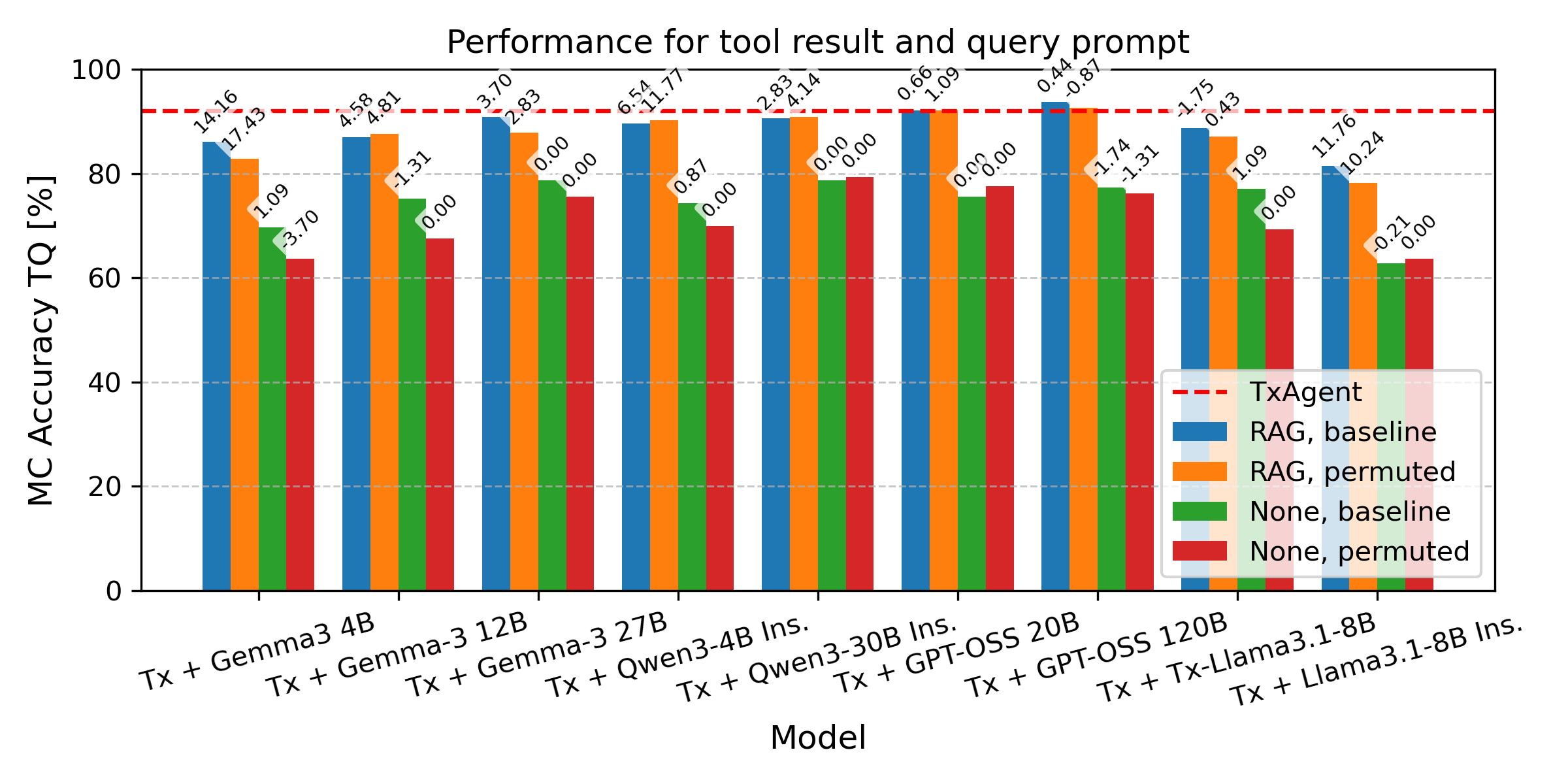}
    \label{fig:acc_mc}
\end{figure}

\subsubsection{Fixed Retrieval Setup}
To investigate the capabilities of various LLMs, we conducted a series of experiments, visualized in Figure~\ref{fig:acc_oe} and Figure~\ref{fig:acc_mc}. As in previous sections, we distinguish between the OE-MC and MC subsets of the validation set. Additionally, we employ a simple prompt structure that provides context to the LLM: the prompt consists first of the retrieved information, followed by the query. This tool-query (TQ) prompt style differs from TxAgent’s prompt, which additionally uses a specialized structure to guide the LLM through the agentic workflow. This structure was removed here, as it gradually reduced the performance of other LLMs.

For each setting, identical retrieved information was used from a sample run of TxAgent. The extracted tool-call context was cleaned to provide a consistent input across all tested LLMs. Furthermore, the answers for OE-MC and MC questions were permuted to maximize the difference from their original positions. Specifically, the answer permutation $[A, B, C, D] \rightarrow [B, D, A, C]$ was applied.

\subsubsection{Fixed Retrieval Results}

For both datasets (OE-MC and MC), we observe that the absence of additional retrieved information decreases the overall performance of the LLMs. Additionally, permuting the answer options generally lowers accuracy across experiments. While there are occasional increases in performance due to permutation, these gains are smaller in magnitude than the decreases. MC question styles are generally easier to solve, as the presence of answer options at the beginning of the iterative answer-generation process allows the LLM to focus its retrieval on relevant information for the suggested answers.  

Compared to the TxAgent baseline, only the GPT-OSS models achieved higher accuracy, highlighting their strong parametric knowledge and their ability to effectively leverage contextual information. This is particularly evident in the no-retrieval setting, where TxAgent outperforms GPT-OSS in the absence of retrieved information. Furthermore, fine-tuning enhances an LLM’s ability to leverage its context, as demonstrated by the higher performance of TxAgent's fine-tuned Llama3.1-8B compared to the non-fine-tuned Llama3.1-8B using identical context.  

It is also noteworthy that smaller models, such as Gemma3-4B and Qwen3-4B, achieve high accuracy in the presence of retrieved information. This indicates that even smaller models can effectively utilize their context windows and could serve as cost-effective alternatives to proprietary or resource-intensive LLMs for RAG settings in therapeutic medical reasoning.

\section{Conclusion and Future Work}\label{sec:conclusion}
In this paper, we describe our work for the CURE-Bench NeurIPS 2025 competition on therapeutic medical agentic question answering, for which we received the \textit{Excellence Award in Open Science}. We outline the competition setup, summarize relevant work on retrieval-augmented generation and tool-calling for integrating up-to-date information, and discuss the limitations of the existing agentic medical reasoning system, TxAgent. We then detail our improvements to this agent. Finally, we present a set of experiments demonstrating the benefits of incorporating an additional source of current medical knowledge and emphasizing the importance of reliable retrieval techniques for function-calling.
\\\\
\textbf{Limitations.}
TxAgent’s retrieval system performs tool-calling with highly precise tool selection to minimize context expansion, resulting in multiple iterative function-call executions. Our extension using DailyMed does not employ such fine-grained selection. Consequently, it may retrieve information that would otherwise require several ToolUniverse function calls, potentially leading to larger context windows and increased computational overhead.

%\input{latex/06_appendix} use "\appendix"?

%\bibliography{custom}

%\appendix

%\section{Example Appendix}
%\label{sec:appendix}

%This is an appendix.

\end{document}